\documentclass{article}
\pdfoutput=1
\usepackage{amsmath,epsfig}
\usepackage[preprint]{spconfa4}
\usepackage{algorithm}
\usepackage{algorithmic}
\usepackage{multirow}
\usepackage{color}
\usepackage{cite}
\usepackage{array}
\usepackage{amssymb} 
\usepackage{color}

\copyrightnotice{978-1-6654-3864-3/21/\$31.00 ©2021 IEEE}

\let\OLDthebibliography\thebibliography
\renewcommand\thebibliography[1]{
	\OLDthebibliography{#1}
	\setlength{\parskip}{0pt}
	\setlength{\itemsep}{0pt plus 0.3ex}
}

\begin{document}\sloppy

	% Example definitions.
	% --------------------
	\def\x{{\mathbf x}}
	\def\L{{\cal L}}

	% Title.
	% ------
	\title{Unsupervised Video Person Re-identification via Noise and Hard frame Aware Clustering}
	%
	% Address.
	% ---------------
	\name{Pengyu Xie${^1}$, Xin Xu${^{1,*}}$\thanks{* Corresponding Author}, Zheng Wang${^{2,3}}$, Toshihiko Yamasaki${^{2,3}}$}
	\address{${^1}$School of Computer Science and Technology, 
		Wuhan University of Science and Technology \\
		${^2}$Research Institute for an Inclusive Society through Engineering (RIISE), The University of Tokyo \\ 
		${^3}$Department of Information and Communication Engineering, The University of Tokyo}

	\maketitle

	\begin{abstract}
		Unsupervised video-based person re-identification (re-ID) methods extract richer features from video tracklets than image-based ones. The state-of-the-art methods utilize clustering to obtain pseudo-labels and train the models iteratively. However, they underestimate the influence of two kinds of frames in the tracklet: 1) noise frames caused by detection errors or heavy occlusions exist in the tracklet, which may be allocated with unreliable labels during clustering; 2) the tracklet also contains hard frames caused by pose changes or partial occlusions, which are difficult to distinguish but informative. This paper proposes a Noise and Hard frame Aware Clustering (NHAC) method. NHAC consists of a graph trimming module and a node re-sampling module. The graph trimming module obtains stable graphs by removing noise frame nodes to improve the clustering accuracy. The node re-sampling module enhances the training of hard frame nodes to learn rich tracklet information. Experiments conducted on two video-based datasets demonstrate the effectiveness of the proposed NHAC under the unsupervised re-ID setting.
	\end{abstract}
	\section{Introduction}
	\label{sec:intro}
	
	Compared with image-based person re-identification (re-ID), video-based person re-ID has gained increasing attention due to the rich spatial-temporal information that exists in video tracklets~\cite{huang2018video}. However, most video-based person re-ID methods are conducted in a supervised manner, requiring intensive manual labeling~\cite{wang2020re}. To be scalable in real-world applications, recent studies focusing on the unsupervised video-based person re-ID~\cite{xu2021rethinking} are broadly divided into two categories.
	
	\begin{figure}[t]
		\centering
		\includegraphics[scale=0.35]{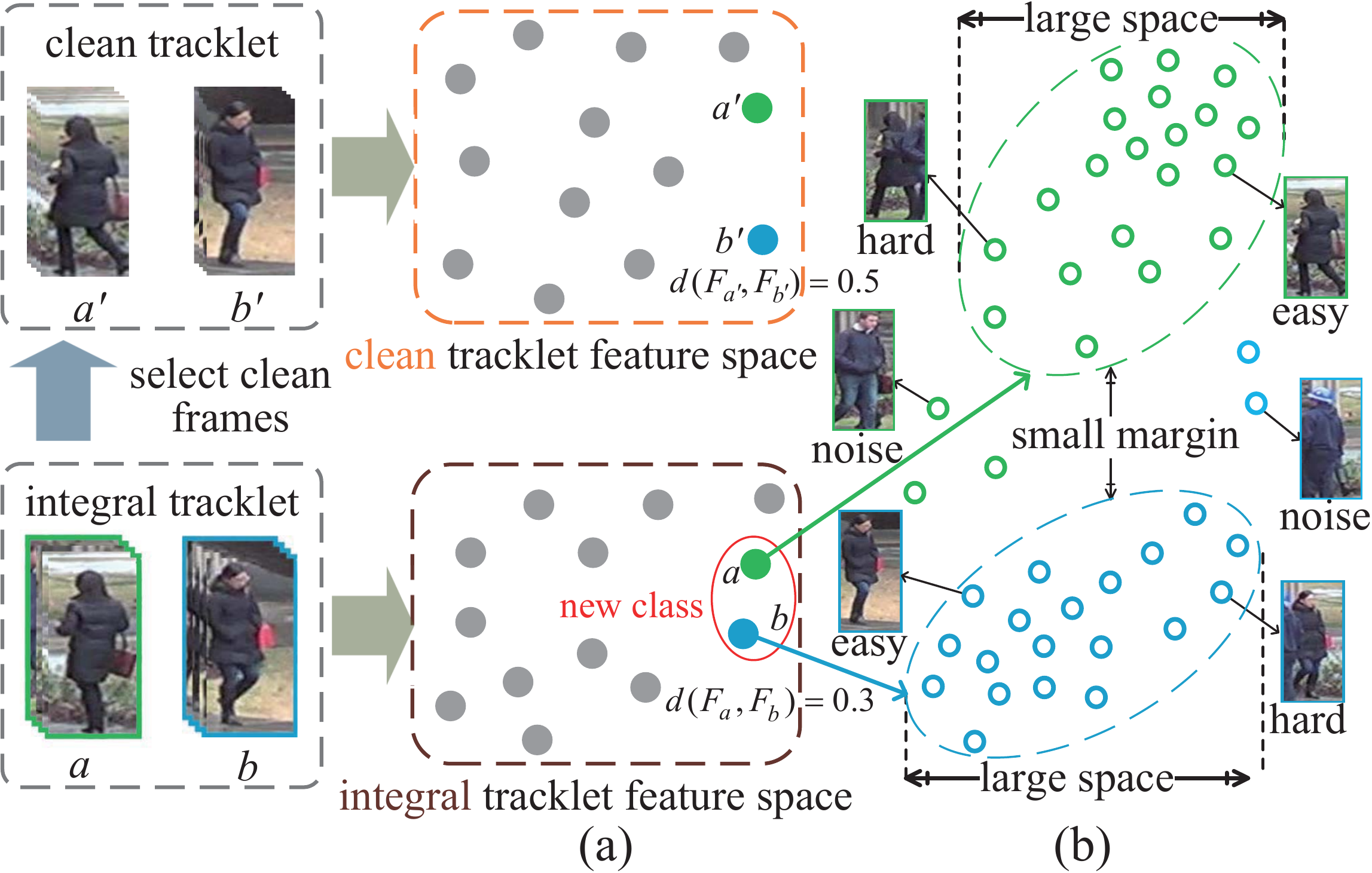}
		\vspace{-2mm}
		\caption{Illustration of the impact of noise and hard frames. 
			(a) shows the feature space of integral and clean tracklet features, respectively. Due to the noise frames, the features of tracklets $a$ and $b$ have a high similarity, and they will be wrongly merged during clustering. When noise frames are trimmed, the features of clean tracklets $a'$ and $b'$ are able to be easily distinguished. 
			(b) shows the feature space of frames in the tracklet. The hard frames are more scattered than easy frames and do not have the same high aggregation frame feature spaces as easy frames. 
		}
		\vspace{-4mm}
		\label{fig:resone}
	\end{figure}
	
	1) Association-based: %the first category utilized 
	methods utilize tracklet association to mine labels from the unlabeled tracklets. Some methods obtain labels under the assumption of long tracklet or spatio-temporal topology~\cite{wu2018exploit,ye2017dynamic,liu2017stepwise}. For example, Ye \textit{et al.}~\cite{ye2017dynamic} employed a static strategy to associate similar tracklet data for further training. Liu \textit{et al.}~\cite{liu2017stepwise} carried out a reciprocal nearest neighbor search for negative sample mining to improve the accuracy of tracklet association. However, these methods still require several pedestrian tracklet labels. To avoid the usage of person identity labeling, other methods explore the tracklet association under different cameras~\cite{li2020unsupervised,chen2018deep}. Li \textit{et al.}~\cite{li2020unsupervised} and Chen \textit{et al.}~\cite{chen2018deep} utilized intra-camera and cross-camera anchors to enhance tracklet association learning. Yet, these methods may face great challenges when camera-related information is not provided.
	
	2) Clustering-based: %the second category utilized
	methods utilize clustering to perform unsupervised learning. These methods generally include two main steps: i) clustering is conducted on the features extracted from the person representation model, then pseudo labels are assigned to each sample according to their clustering results; ii) the model is retrained with samples and their pseudo labels. Lin \textit{et al.}~\cite{lin2019bottom} proposed a Bottom-Up Clustering (BUC) method to enhance the intra-cluster tightness and inter-cluster well-separation. Ding \textit{et al.}~\cite{ding2019dispersion} proposed a Dispersion Based Clustering (DBC) method to build high-quality clusters. Recently, Wu \textit{et al.}~\cite{wu2020tracklet} took full advantage of the intrinsic tracklet appearance information to formulate a novel Tracklet Self-Supervised Learning (TSSL) method.

	However, as shown in Figure~\ref{fig:resone}, current clustering-based methods underestimate the influence of two kinds of frames in the tracklet. 1) Noise frames: the tracklet may contain noise frames caused by detection errors or heavy occlusions. These noise frames introduce certain bias to the tracklet feature (the tracklet feature is usually represented by average pooling to multiple frame features). As a result, different pedestrians may share a high similarity illustrated in Figure~\ref{fig:resone} (a), $a$ and $b$ are two biased tracklets with different IDs but are easily merged during clustering. 
	2) Hard frames: there are hard frames in the tracklet caused by pose changes or partial occlusions of the pedestrian. These hard frames are difficult to distinguish but contain rich information. As shown in Figure~\ref{fig:resone} (b), the feature distribution of hard frames is more scattered than that of easy frames in the frame feature spaces. As a result, the features within the same tracklets are not sufficiently close, and the margins between different tracklets are not adequately large.
	
	Aiming to address these problems, this paper proposes a Noise and Hard frame Aware Clustering (NHAC) method to optimize clustering accuracy and feature embedding space for unsupervised video-based person re-ID. To suppress the feature bias caused by noise nodes, a graph trimming module is presented to improve the accuracy of tracklet similarity. The graph trimming module utilizes the diversities between features to identify and remove noise nodes. Then, a stable tracklet graph structure is obtained to make the characteristics of the tracklet more accurate. To learn the rich information from hard frame nodes, a node re-sampling module is designed to learn abundant pedestrian tracklet information and obtain a discriminative feature space. This module explores the dispersion of tracklets to identify hard nodes and to learn the rich information of tracklets between the input data and feature embedding levels. Our contributions in this paper are summarized in two folds:
	\vspace{-1mm}
	\begin{itemize}
		
		\item We propose a noise and hard frame aware clustering method for unsupervised video-based person re-ID. It is built on the application-specific characteristics existing in the video re-ID task.
		\vspace{-1mm}
		\item We design a graph trimming module to avoid feature bias caused by noise frames and design a node re-sampling module to enhance the training of hard frames.
	\end{itemize}
	
	\section{Methodology}
\begin{figure*}[t]
		\centering
		\includegraphics[width=0.85\textwidth]{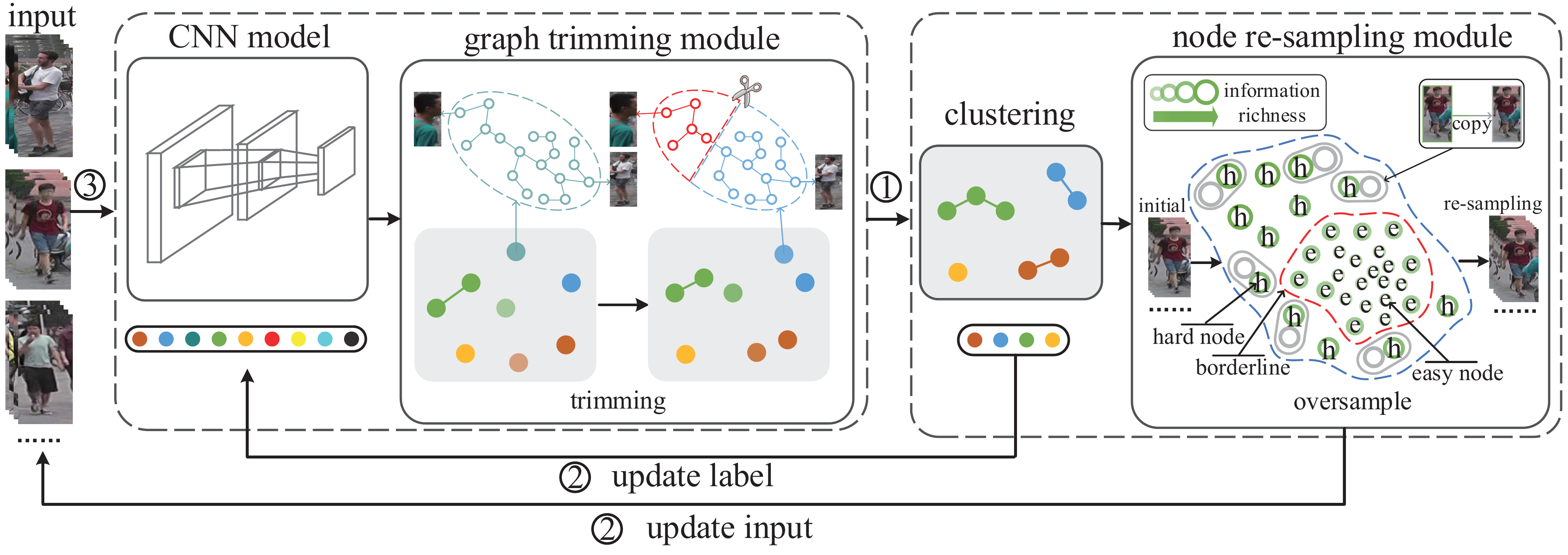}
		\vspace{-4mm}
		\caption{The framework of our method. The CNN model is initially trained with the initial pseudo labels. In each cycle, tracklet feature embeddings are firstly extracted by the graph trimming module; after clustering, new training data are obtained through the node re-sampling module; then, the network is re-trained with new pseudo labels and input data.}
		\label{fig:restwo}
		\vspace{-4mm}
	\end{figure*}

	The proposed NHAC consists of three main parts: 1) A clustering framework is utilized to obtain a reliable cluster structure. 2) The graph trimming module improves the clustering accuracy by removing the noise frames node. 3) The node re-sampling module enhances the training of tracklet hard frames node to learn abundant pedestrian tracklet information. Figure~\ref{fig:restwo} illustrates the overview of NHAC.
	
	\vspace{-2mm}
	\subsection{The Clustering Framework}
	Clustering is an effective strategy for unsupervised learning. However, it is non-trivial to straightforwardly utilize clustering for person re-ID, since person re-ID is a fine-grained recognition task with large intra-class differences whilst subtle inter-class variations properties. To obtain a suitable cluster structure for person re-ID, we employ the BUC clustering-based unsupervised re-ID method proposed by Lin \textit{et al.}~\cite{lin2019bottom}. This method iteratively updates the CNN by two steps, \textit{i.e.} model training and cluster merging.
	
	Given an unlabeled training set $\chi  = \{ {X_1},{X_2}, \ldots ,{X_N}\} $ of $N$ tracklets, each tarcklet contains $L$ frames, \textit{i.e.} ${X_i} = \{ x_i^j\} _{j = 1}^L$. To explore the clustering of tracklet, we deploy a feature embedding model ${f_\theta }( \cdot )$ (where $\theta$ is the network parameters). During the training, we randomly select $M$ frames in the tracklet to extract features, and then average pooling these features. It is formulated as $\boldsymbol{v}_i = \rm{avgpool}({f_\theta }( X_{i}^M ))$. Since we do not have ground truth labels in the initial phase, we allocate each tracklet to a different cluster, so each tracklet has a pseudo label. The tarcklet $X_i$ belongs to the $c$-th cluster probability is defined as:
	\begin{equation}
	    \setlength{\abovedisplayskip}{2pt}
        \setlength{\belowdisplayskip}{2pt}
		\label{eq_one}
		p(c|X_i,V) = \frac{{\exp (\boldsymbol{V}_{c}^{\rm{T}}\boldsymbol{v}_i/\tau )}}{{\sum\nolimits_{j = 1}^C {\exp (\boldsymbol{V}_{j}^{\rm{T}}\boldsymbol{v}_i/\tau )} }},
	\end{equation}
	where $\boldsymbol{V} \in { R^{{n_\theta } \times C }}$ is the lookup table that stores the centroid feature of each cluster, $n_\theta$ is the dimension of the feature, $C$ is the number of clusters at the present stage, ${\boldsymbol{V}_j}$ is the $j$-th column of $\boldsymbol{V}$ and $\tau $ is a temperature parameter. At the initial training stage, $C$ is the number of tarcklets and at the following stages, $C$ will gradually decrease. During the training, we calculate cosine similarities between feature ${\boldsymbol{v}_i}$ with all the centroid feature of each cluster by ${\boldsymbol{V}^{\rm{T}}} \cdot {\boldsymbol{v}_i}$. Then we update the $y_i$-th column of table $\boldsymbol{V}$ by ${\boldsymbol{V}_{{y_i}}} = \frac{1}{2}({\boldsymbol{V}_{{y_i}}} + {\boldsymbol{v}_i})$. Finally, the loss function is formulated as:
	\begin{equation}
	    \setlength{\abovedisplayskip}{2pt}
        \setlength{\belowdisplayskip}{2pt}
		\label{eq_two}
		{\cal L }{\rm{ =  - log(p(}}{{y}_i}{|}{{X}_i}{,\boldsymbol{V}))}.
	\end{equation}
	
	The cluster merging is vital in each iteration of the clustering. Following Lin \textit{et al.}~\cite{lin2019bottom}, the distance between two clusters is represented by the minimum pairwise distance. The trained model ${f_\theta }( \cdot )$ is used to extract tracklet features $\beta = \{{F_1},{F_2}, \ldots ,F_N \}$. The distance between cluster $A$ and $B$ is formulated as:
	${D}(A,B) = \mathop {\min }\limits_{{a} \in A,{b} \in B} d({F_a},{F_b})$. ${F_a},{F_b}$ are the features of the tracklets in clusters $A$ and $B$ respectively. $d({F_a}$, ${F_b})$ is defined as the euclidean distance between the feature embeddings of two tracklets. In order to make the merged clusters more reliable, the method is based on a hierarchical clustering algorithm. In each iteration of clustering, this method selects the nearest top-$m$ clusters to merge, where $m = N * mp$, and $mp \in (0,1)$ denotes the merging percentage of clustering. Then, according to the result of cluster merging to assign pseudo labels to each tracklet. In the end, this framework iteratively updates the CNN to enhance model performance through the above two steps.
	\vspace{-4mm}
	\subsection{Graph Trimming Module}
	In video-based person re-ID, each tracklet is usually automatically acquired by the detector. However, the tracklet may contain noise frames caused by detection errors. In addition, heavy occlusions of pedestrians may also generate noise frames. Because the tracklet features are usually represented by average pooling to transform multiple frame vectors into a single feature vector, noise frames may easily bias the tracklet feature. Thus, we design a graph trimming module to reduce the impact of noise frames. 
	
	For a single tracklet, we treat each frame as a node, and multiple nodes constitute a tracklet graph. Compared with other nodes in the tracklet graph, we find a large discrepancy in the noise nodes. Specifically, for each tracklet ${X_i} = \{ x_i^j\} _{j = 1}^L$, the features of all nodes are extracted as $\hat{F_i} = \{ f_i^j\} _{j = 1}^L$. Then we use average pooling to get the graph central node feature ${F_i} = \mathrm{avgpool}(\hat{F_i})$. We calculate each node with central node cosine similarity $s_i^j = \cos (f_i^j,{F_i})$. The cosine similarity between $\hat{F_i}$ and $F_i$ is expressed as ${S_i} = \{s_i^1,s_i^2, \ldots ,s_i^L\} $. Due to the large bias of noise nodes, intuitively, we can trim some noise nodes by setting a fixed threshold according to the similarity.
	
	However, this method presents two problems. 1) There are variations between different datasets, and their corresponding threshold should be different. 2) Different tracklets of the same dataset have different lengths and qualities, it is easy to make the judgment of noise nodes inaccurate if the threshold is fixed. In order to trim accurately more noise nodes, we set the dynamic threshold by calculating the variance of the graph. Specifically, the dynamic threshold ${q_i}$ is expressed as:
	\begin{equation}
	    \setlength{\abovedisplayskip}{2pt}
        \setlength{\belowdisplayskip}{2pt}
		\label{eq_three}
		{q_i} = {{\sum\nolimits_{j = 1}^L {u_i^j} } \mathord{\left/
				{\vphantom {{\sum\nolimits_{j = 1}^L {u_i^j} } {(L*\delta )}}} \right.
				\kern-\nulldelimiterspace} {(L*\delta )}},
	\end{equation}
	where $L$ is the number of nodes in the tracklet graph, and $\delta$ is the parameter that controls the degree of noise relaxation. $u_i^j = {(1 - s_i^j)^2}$ is the square of the inverse cosine similarity between the current node and the center node. For node $x_i^j$, when $u_i^j > {q_i}$, it is judged as a noise node and needs to be trimmed. For each tracklet, after graph trimming node, the reserved features are expressed as $\hat{F_{i^{'}}}=\{f_{i}^{j}|u_{i}^{j}<q_i,j \in \{ 1,2, \ldots ,L\}\}$ and the trimmed tracklet feature is expressed as ${F_{i^{'}}} = \mathrm{avgpool}({\hat{F_{i^{'}}}})$. Finally, through the graph trimming module, the distance between cluster $A$ and $B$ is formulated as ${D}(A,B) = \mathop {\min }\limits_{{{a^{'}}} \in A,{{b^{'}}} \in B} d({F_{a^{'}}},{F_{b^{'}}})$.
	
	\tabcolsep=9pt
	\begin{table*}[t]
		\centering
		\caption{Comparisons with state-of-the-arts. ``One'', ``Camera'', and ``None'' denote one-example annotation, camera annotation, and no extra annotation, respectively. "-" denotes that the results are not provided in the original paper. $\rm{1^{st}/2^{nd}}$ best results are in \textbf{bold}$\rm{/}$\underline{underline}.
		}
		\vspace{0.05cm}
		\label{tb1}
		\resizebox{0.89\textwidth}{!}{
			\begin{tabular}{l|c|c|cccc|cccc}
				\hline
				\multirow{2}{*}{Methods} & \multirow{2}{*}{Venue} & \multirow{2}{*}{Annotation} & \multicolumn{4}{c|}{DukeMTMC-VideoReID} & \multicolumn{4}{c}{MARS}        \\ \cline{4-11} 
				&                        &                         & Rank-1   & Rank-5   & Rank-10   & mAP   & Rank-1 & Rank-5 & Rank-10 & mAP  \\ \hline
				
				DGM+IDE~\cite{ye2017dynamic}                  & ICCV'17                 & One                    & 42.3     & 57.9     & 68.3      & 33.6  & 36.8   & 54.0     & -       & 16.8 \\
				Stepwise~\cite{liu2017stepwise}                 & ICCV'17                 & One                    & 56.2     & 70.3     & 79.2      & 46.7  & 41.2   & 55.5   & -       & 19.6 \\
				RACE~\cite{ye2018robust}                     & ECCV'18                 & One                    & -        & -        & -         & -     & 43.2   & 57.1   & 62.1    & 24.5 \\
				EUG~\cite{wu2018exploit}                      & CVPR'18                 & One                    & 72.7     & 84.1     & -         & 63.2  & \textbf{62.6}   & \underline{74.9}   & -       & \textbf{42.4} \\
				DAL~\cite{chen2018deep}                      & BMVC'18                 & Camera                  & -        & -        & -         & -     & 49.3   & 65.9   & 72.2    & 23   \\
				UTAL~\cite{li2020unsupervised}                      & TPAMI'20                 & Camera                  & -        & -        & -         & -     & 49.9   & 66.4   & -    
				& 35.2 \\
				OIM~\cite{xiao2017joint}                      & CVPR'17                 & \textbf{None}           & 51.1     & 70.5     & 76.2      & 43.8  & 33.7   & 48.1   & 54.8    & 13.5 \\
				BUC~\cite{lin2019bottom}                      & AAAI'19                 & \textbf{None}           & 74.8     & 86.8     & 89.7      & 66.7  & 55.1   & 68.3   & 72.8    & 29.4 \\
				DBC~\cite{ding2019dispersion}                 & BMVC'19                 & \textbf{None}           & \underline{75.6}     & \underline{88.5}     & \underline{91.0}      & \underline{67.4}  & 58.5   & 70.1   & \underline{73.5}
				& 31.7 \\
				TSSL~\cite{wu2020tracklet}                     & AAAI'20                 & \textbf{None}           & 73.9     & -        & -         & 64.6  & 56.3   & -      & -       & 30.5 \\ \hline
				\textbf{Ours}            & -                      & \textbf{None}           & \textbf{82.8}       & \textbf{92.7}       & \textbf{95.6}        & \textbf{76.0}    & \underline{61.8}     & \textbf{75.3}     & \textbf{79.9}      & \underline{40.1}   \\ \hline
			\end{tabular}
		}
		\vspace{-0.4cm}
	\end{table*}
	\vspace{-4mm}
	\subsection{Node Re-sampling Module}
	In order to make the model learn rich information from hard nodes of the tracklet during training, we propose a node re-sampling module to improve the discriminability of the tracklet feature space.
	
	For node $\{ x_i^j\} _{j = 1}^L$ in tracklet ${X_i}$, the less information will be learned if it is easier to be distinguished. To find nodes with rich information, the predicted probability of the current node ID is usually utilized for judgment. However, it is impossible to judge the correctness of the pseudo label of the current tracklet for unsupervised tasks. For each node in the tracklet, the closer it is to the central node, the less information is obtained by the model training. Similar to the graph trimming module, for each tracklet, we calculate the cosine similarity between each node with the central node, \textit{i.e.} ${S_i} = \{s_i^1,s_i^2, \ldots ,s_i^L\}$, and calculate its mean, \textit{i.e.} ${\bar{s_i}} = {{\sum\nolimits_{j = 1}^L {s_i^j} } \mathord{\left/	{\vphantom {{\sum\nolimits_{j = 1}^L {s_i^j} } L}} \right.\kern-\nulldelimiterspace} L}$.
	According to the value ${\bar{s_i}}$, the node is divided into two sets ${g_i},{b_i}$. When $s_i^j > \bar{s_i}$, the node is easily distinguished, \textit{i.e.} $g_{i}=\{x_{i}^{j}|s_{i}^{j}>\bar{s_i},j \in \{ 1,2, \ldots ,L\}\}$; otherwise the node is hard to be distinguished, \textit{i.e.} $b_{i}=\{x_{i}^{j}|s_{i}^{j}<\bar{s_i},j \in \{ 1,2, \ldots ,L\}\}$. We found that the number of ${g_i}$ nodes in a tracklet is usually more than the number of ${b_i}$ nodes. In order to learn more information from hard nodes during training, the set ${b_i}$ is oversampled to the length of set ${g_i}$, \textit{i.e.} $b_i^* = \mathrm{oversampling}{({b_i})_{\mathrm{len}({g_i})}}$. Finally, we get a new set ${E_i}$ through oversampling, $M$ frame nodes in set $E_i$ is randomly selected during training.
	\begin{equation}
	    \setlength{\abovedisplayskip}{2pt}
        \setlength{\belowdisplayskip}{2pt}
		\label{eq_four}
		{E_i} = \left\{ {
			\begin{array}{*{20}{l}}
				{[{g_i},b_i^*],\qquad \rm{if} \  \mathrm{len}({g_i}) > \mathrm{len}({b_i})}\\
				{[{g_i},{b_i}],\qquad \rm{others}}
		\end{array}} 
		\right\}.
	\end{equation}
	
	In addition, the undersampling and its combination with oversampling also enhance the learning of hard nodes. 1) The undersampling method randomly selects $n$ nodes in $g_i$ to obtain set $g_i^*$, where $n$ is the length of $b_i$. Then $g_i^*$ is combined with $b_i$ to get a new train set. However, random selection is uncertain and may reduce the diversity of nodes available for training. 2) The oversampling and undersampling union method combines the sets $b_i^*$ and $g_i^*$ to get a new train set. Although this method improves the learning of hard nodes, it may lead the model to overfit the minority samples. In experiments, the effectiveness of the oversampling method will also be demonstrated in Section \ref{DS}.
	
	Although the model has learned rich information from hard nodes in the training data, the distribution of the feature space still needs to be improved. To directly optimize feature space, triplet loss is utilized in this work which needs to generate batches of positive and negative pairs during training. The characteristics of the video tracklet are used to design tracklet triplet loss. In the same tracklet, the labels of different nodes should be consistent. Therefore, the positive pairs are easy to be formed between nodes in the same tracklet. During training, $M$ frames are randomly selected to get set ${X_{i}^{M}}$ in each tracklet, which is divided into $K$ parts with length $M/K$. Since these $K$ parts have the same label, the anchor $x_i$ and the positive pairs $x_{i,p}^*$  are easy to be obtained. For negative pairs, pseudo labels are used to generate negative pairs $x_{i,n}^*$. The proposed tracklet triplet loss is expressed as:
	\begin{equation}
	    \setlength{\abovedisplayskip}{2pt}
        \setlength{\belowdisplayskip}{2pt}
		\label{eq_five}
		{{\cal L}_t} = \max (0,\alpha  + D({v_{{x_i}}},{v_{x_{i,p}^*}}) - D({v_{{x_i}}},{v_{{x_{i,n}^*}}})),
	\end{equation}
	where $\alpha$  denotes a margin, $\{{v_{x_i}},v_{x_{i,p}^*},{v_{x_{i,n}^*}}\}$ are the feature vectors of $\{ {x_i},x_{i,p}^*,{x_{i,n}^*}\} $ respectively, $x_{i,p}^*$ is positive tracklet and ${x_{i,n}^*}$  is negative tracklet. 
	\vspace{-2mm}
	\section{Experiments}
	\vspace{-2mm}
	\subsection{Datasets}
	\vspace{-2mm}
	We conduct extensive experiments on the DukeMTMC-VideoReID~\cite{wu2018exploit} dataset and the MARS~\cite{zheng2016mars} dataset to evaluate the proposed NHAC. On both the DukeMTMC-VideoReID and the MARS, each tracklet is treated as an individual sample in the model training. Note that, our method does not utilize any annotation information for model initialization or training. We adopt Rank-$k$ and mean average precision (mAP) to evaluate our method.
	\vspace{-3mm}
	\subsection{Implementation Details}
	\vspace{-1mm}
	We adopt ResNet-50 as the CNN backbone to conduct all the experiments and initialized it using the ImageNet pre-trained model. During training, the number of training epochs in the first stage, the batch size, and the dropout rate were set to be 20, 16, and 0.5 respectively. The parameters $mp$, $K$, and $\delta $ in  Eq.(\ref{eq_three}) were set to 0.05, 2, and 0.5 respectively. $\alpha$ in Eq.(\ref{eq_five}) is set to 0.3. We set $\tau $ = 0.1 following~\cite{xiao2017joint} and select $M=16$ frames as the input for each tracklet. In the cluster merging and final evaluation, average pooling was used to aggregate the frame-level features into a tracklet representation. We adopt stochastic gradient descent (SGD) with a momentum of 0.9 to optimize the model. The learning rate is initialized to 0.1 and set to 0.01 after 15 epochs.
	We fixed the first three residual blocks of ResNet-50 to save GPU memory and to boost iterations. The proposed method was implemented on Pytorch and trained with NVIDIA TITAN Xp GPU.

	\vspace{-2mm}
	\subsection{Comparison with the State-of-the-Arts}
	\vspace{-1mm}
	In Table \ref{tb1}, we compare our method with the state-of-the-art algorithms on the two large-scale video-based datasets. On the DukeMTMC-VideoReID dataset, our NHAC achieves 76.0\% in mAP and 82.8\% in Rank-1, improving the state-of-the-art performance by 8.6 points and 7.2 points, respectively. Compared to the BUC method, our method achieves 9.3 and 8.0 points of improvement on mAP accuracy and Rank-1, respectively.
	
	On the MARS dataset, we achieve 40.1\% in mAP and 61.8\% in Rank-1, which beats the state-of-the-art cluster-based algorithms by 8.4 and 3.3 points, respectively. Compared to BUC, we achieve 10.7 and 6.7 points of improvement in mAP accuracy and Rank-1. Compared with the association-based approaches, our method outperforms most existing state-of-the-art methods except EUG. As stated in ~\cite{wu2018exploit}, EUG initializes its model with a small number of tracklet labels. Therefore, EUG is not fully unsupervised. Compared with UTAL and DAL which utilize camera annotations, our method gives the best performance.

	\begin{table}[t]
		\centering
		\vspace{-0.2cm}
		\caption{Evaluating the components of NHAC on the DukeMTMC-VideoReID and the MARS datasets.}
		\vspace{0.1cm}
		\label{tb2}
		\resizebox{0.8\linewidth}{!}{
			\begin{tabular}{l|cc|cc}
				\hline
				\multirow{2}{*}{Components} & \multicolumn{2}{c|}{DukeMTMC} & \multicolumn{2}{c}{MARS} \\ \cline{2-5} 
				& Rank-1              & mAP               & Rank-1       & mAP        \\ \hline
				Baseline                    & 74.5                & 66.8              & 56.1         & 31.6       \\ \hline
				NHAC \textit{w/o} NRM       & 76.5                & 69.3              & 60.0         & 35.4       \\ \hline
				NHAC \textit{w/o} GTM       & 81.5                & 74.9              & 59.9         & 37.1       \\ \hline
				NHAC                        & 82.8                & 76.0              & 61.8         & 40.1       \\ \hline
			\end{tabular}
		}
		\vspace{-0.3cm}
	\end{table}
	
	\tabcolsep=0.2pt
	\begin{figure}[t]
		\centering
		\begin{tabular}{ccccc ccc cc}
			\includegraphics[width=0.07\linewidth,height=1.3cm]{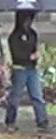} &
			\includegraphics[width=0.07\linewidth,height=1.3cm]{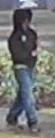} & 
			\includegraphics[width=0.07\linewidth,height=1.3cm]{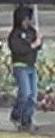} &
			\includegraphics[width=0.07\linewidth,height=1.3cm]{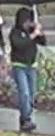} & 
			\includegraphics[width=0.07\linewidth,height=1.3cm]{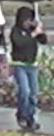} &
			\quad
			\includegraphics[width=0.07\linewidth,height=1.3cm]{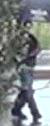} &
			\includegraphics[width=0.07\linewidth,height=1.3cm]{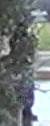}& 
			\quad
			\includegraphics[width=0.07\linewidth,height=1.3cm]{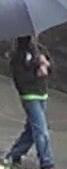} & 
			\includegraphics[width=0.07\linewidth,height=1.3cm]{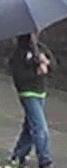} & 
			\includegraphics[width=0.07\linewidth,height=1.3cm]{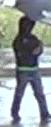}\\
			
			\includegraphics[width=0.07\linewidth,height=1.3cm]{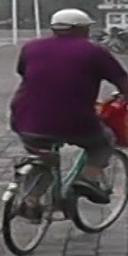} &
			\includegraphics[width=0.07\linewidth,height=1.3cm]{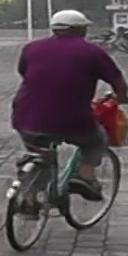} & 
			\includegraphics[width=0.07\linewidth,height=1.3cm]{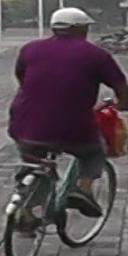} &
			\includegraphics[width=0.07\linewidth,height=1.3cm]{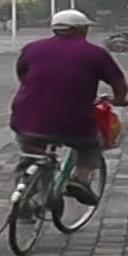} & 
			\includegraphics[width=0.07\linewidth,height=1.3cm]{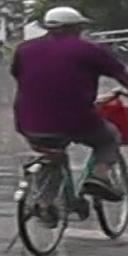} &
			\quad
			\includegraphics[width=0.07\linewidth,height=1.3cm]{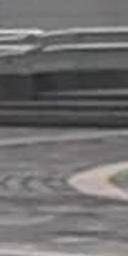} &
			\includegraphics[width=0.07\linewidth,height=1.3cm]{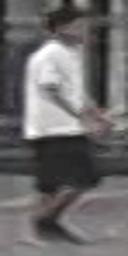}&
			\quad
			\includegraphics[width=0.07\linewidth,height=1.3cm]{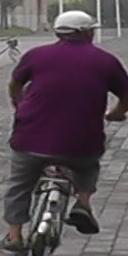} & 
			\includegraphics[width=0.07\linewidth,height=1.3cm]{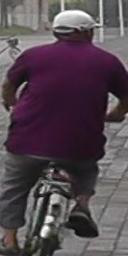} &
			\includegraphics[width=0.07\linewidth,height=1.3cm]{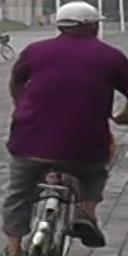}\\
			\multicolumn{5}{c}{\small{easy}}&
			\multicolumn{2}{c}{\small{noise}}&
			\multicolumn{3}{c}{\small{hard}} \\
		\end{tabular}
		\vspace{-4mm}
		\caption{Examples of easy, noise and hard nodes in tracklets.  
		}
		\vspace{-5mm}
		\label{fig:hard} %% label for entire figure
	\end{figure}
	
	\vspace{-4mm}
	
	\subsection{Diagnostic Studies}
	\vspace{-1mm}
	\label{DS}
	To evaluate the effectiveness of our method, we analyze the impact of the two main modules (graph trimming, node re-sampling), hyper-parameter $\delta$, re-sampling criterion, and robustness.
	
	\noindent\textbf{Graph Trimming Module.}
	The Graph Trimming Module (GTM) assigns more accurate pseudo labels during clustering.  As shown in Table \ref{tb2} (see NHAC \textit{w/o} NRM), we show the impact of the GTM on the clustering algorithm. On the DukeMTMC-VideoReID dataset, the module is 2.0\% and 2.5\% higher than baseline in Rank-1 and mAP, respectively. On the MARS dataset, the improvements are 3.9\% for Rank-1 and 3.8\% for mAP. The improvement is more obvious for GTM on the MARS dataset. Because the overall quality of the tracklets in the MARS dataset is higher and the noise frames are more distinct compared to other frames, they can be accurately cropped to achieve good performance. Figure~\ref{fig:hard} illustrates some examples of noise nodes in the tracklets.
	
	\noindent\textbf{Node Re-sampling Module.}
	We evaluate the effectiveness of our Node Re-sampling Module (NRM) by comparing it to our baseline as shown in Table \ref{tb2} (see NHAC \textit{w/o} GTM). NRM helps the model to learn from hard frames at both the input data and feature embedding levels. Therefore, the performance is significantly improved by 7.0\% Rank-1 and 8.1\% mAP on the DukeMTMC-VideoReID, while improving 3.8\% Rank-1 and 5.5\% mAP on the MARS. Note that, on the DukeMTMC-VideoReID dataset, the improvement of this module is more significant. The main reason for this is that on DukeMTMC-VideoReID dataset has a more salient hard frame variance and longer tracklet length compared to MARS (around twice of MARS), which makes the model learn richer information. Some examples of hard nodes are shown in Figure~\ref{fig:hard}.
	
	\begin{figure}[t]
		\centering
		\begin{tabular}{cc}
			\multicolumn{2}{c}{\includegraphics[width=0.96\linewidth]{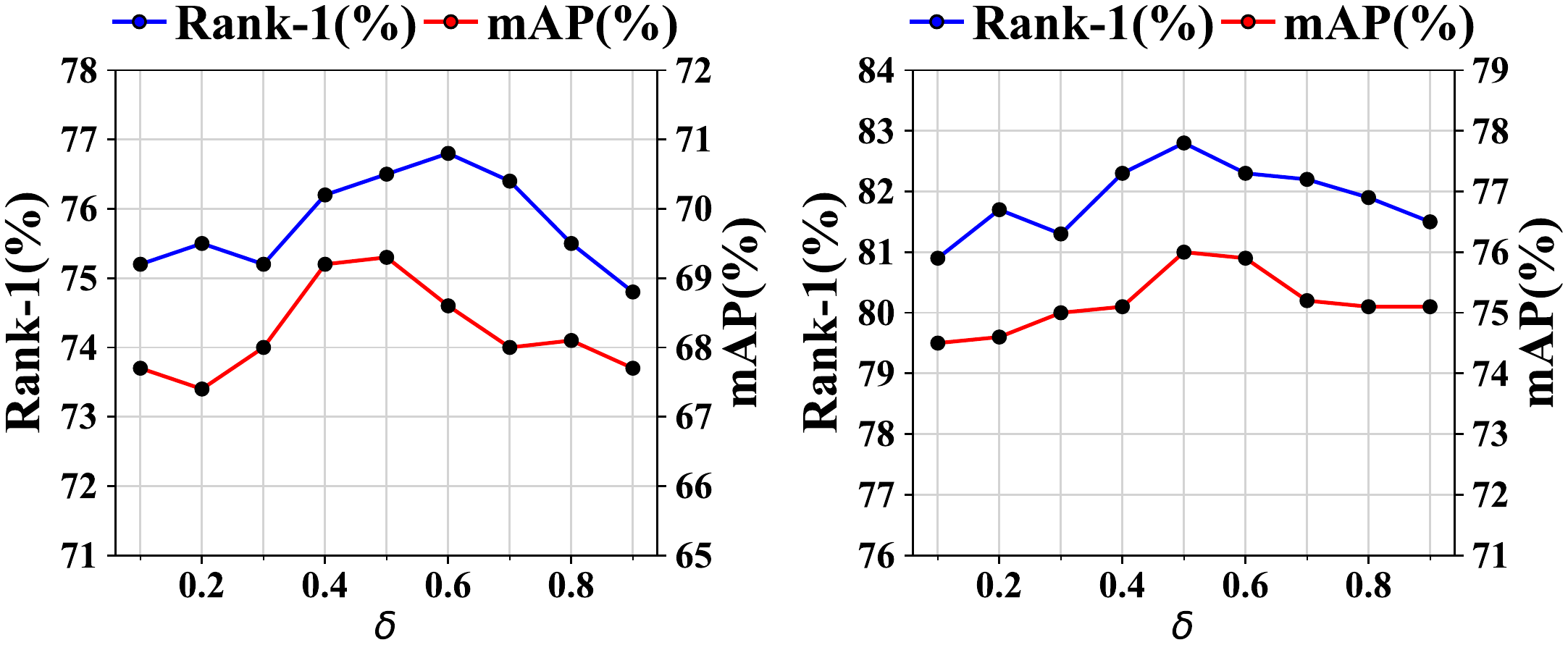}} \\
			\qquad (a) NHAC \textit{w/o} NRM & (b) NHAC        
		\end{tabular}
		\vspace{-0.2cm}
		\caption{Analysis of the parameter $\delta$ on the DukeMTMC-VideoReID dataset. (a) NHAC \textit{w/o} NRM: only have the Graph Trimming Module (GTM); (b) NHAC: the method with GTM and NRM modules.}
		\label{fig:resthree}
		\vspace{-2mm}
	\end{figure}
	
	\noindent\textbf{Hyper-parameter $\delta$.}
	Hyper-parameter $\delta$ controls the degree of identifying noise relaxation, which is helpful to improve the accuracy of trimming noise nodes. We evaluate different values for $\delta$ in Figure~\ref{fig:resthree}. We evaluate the performance of NHAC \textit{w/o} NRM and NHAC, respectively, as the value of $\delta$ varies from 0.1 to 0.9. For NHAC \textit{w/o} NRM, Figure~\ref{fig:resthree} shows that Rank-1 and mAP achieve their highest value when $\delta$=0.6 and $\delta$=0.5, respectively. For NHAC, it consists of modules GTM and NRM. The best Rank-1 and mAP are obtained when $\delta$=0.5. The reason is that when $\delta$ is too small, the degree of identifying noise is relaxing, which makes the module difficult to trim noise nodes. When $\delta$ is too large, a large number of non-noise nodes will be trimmed, which results in insufficient diversity of tracklet features.
	
		\tabcolsep=9pt
	\begin{table}[t]
		\centering
		\vspace{-0.3cm}
		\caption{Comparison of different re-sampling criteria on the DukeMTMC-VideoReID dataset.}
		\vspace{0.1cm}
		\label{tb4}
		\resizebox{0.8\linewidth}{!}{
			\begin{tabular}{l|cccc}
				\hline
				Criterion  & Rank-1 & Rank-5 & Rank-10 & mAP  \\ \hline 
				Over  & \textbf{82.8}   & \textbf{92.7}     & \textbf{95.6}    & \textbf{76.0} \\
				Under & 81.9   & 92.0     & 95.2    & 75.1 \\
				Over+Under & 80.9   & 92.3     & 94.0    & 73.5 \\ \hline
			\end{tabular}
		}
		\vspace{-0.5cm}
	\end{table}
	
	\noindent\textbf{Re-sampling Criterion.}
	Table \ref{tb4} illustrates the results of three re-sampling criteria. The oversampling method (Over) achieves the best result with the Rank-1 = 82.8\% and mAP = 76.0\%. For the undersampling method (Under), we observe a slightly lower performance with the Rank-1 = 81.9\% and mAP = 75.1\%. When using the oversampling and undersampling union method (Over+Under), we observe the Rank-1 and mAP accuracy of 80.9\% and 73.5\%, respectively. The main reason may lie in the different percentages of easy and hard frames in each tracklet, where the oversampling and undersampling union method may lead the model to overfit the minority samples.
	
	\begin{figure}[t]
		\centering
		\begin{tabular}{l}{
				\includegraphics[width=0.85\linewidth]{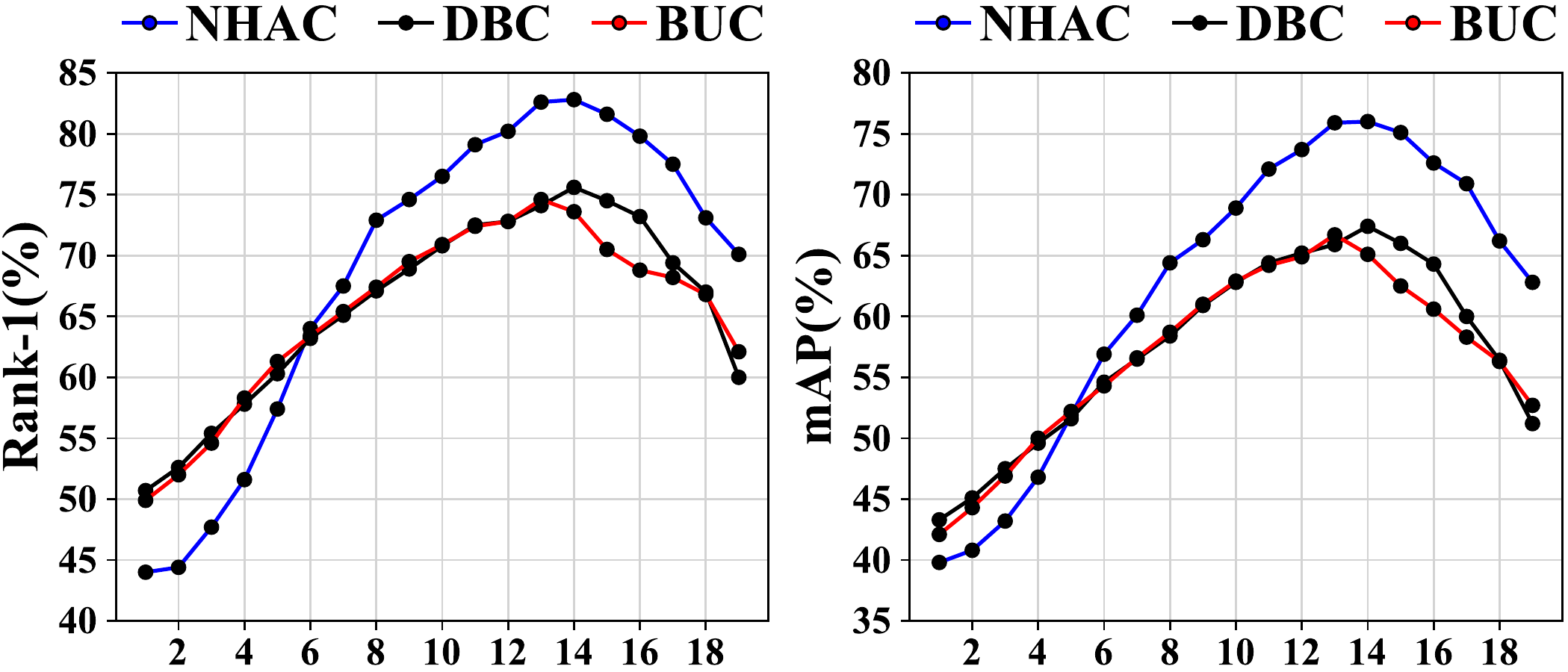}
			}
		\end{tabular}
		\vspace{-4mm}
		\caption{The Rank-1 and mAP performances with the different iterations on DukeMTMC-VideoReID dataset. % Our proposed method (NHAC) have better robustness to the number of iteraions, compared to baseline.
		}
		\label{fig:resfour}
		\vspace{-2mm}
	\end{figure}
	
	\begin{figure}[t]
		\centering
		\begin{tabular}{l}{
				\includegraphics[width=0.58\linewidth]{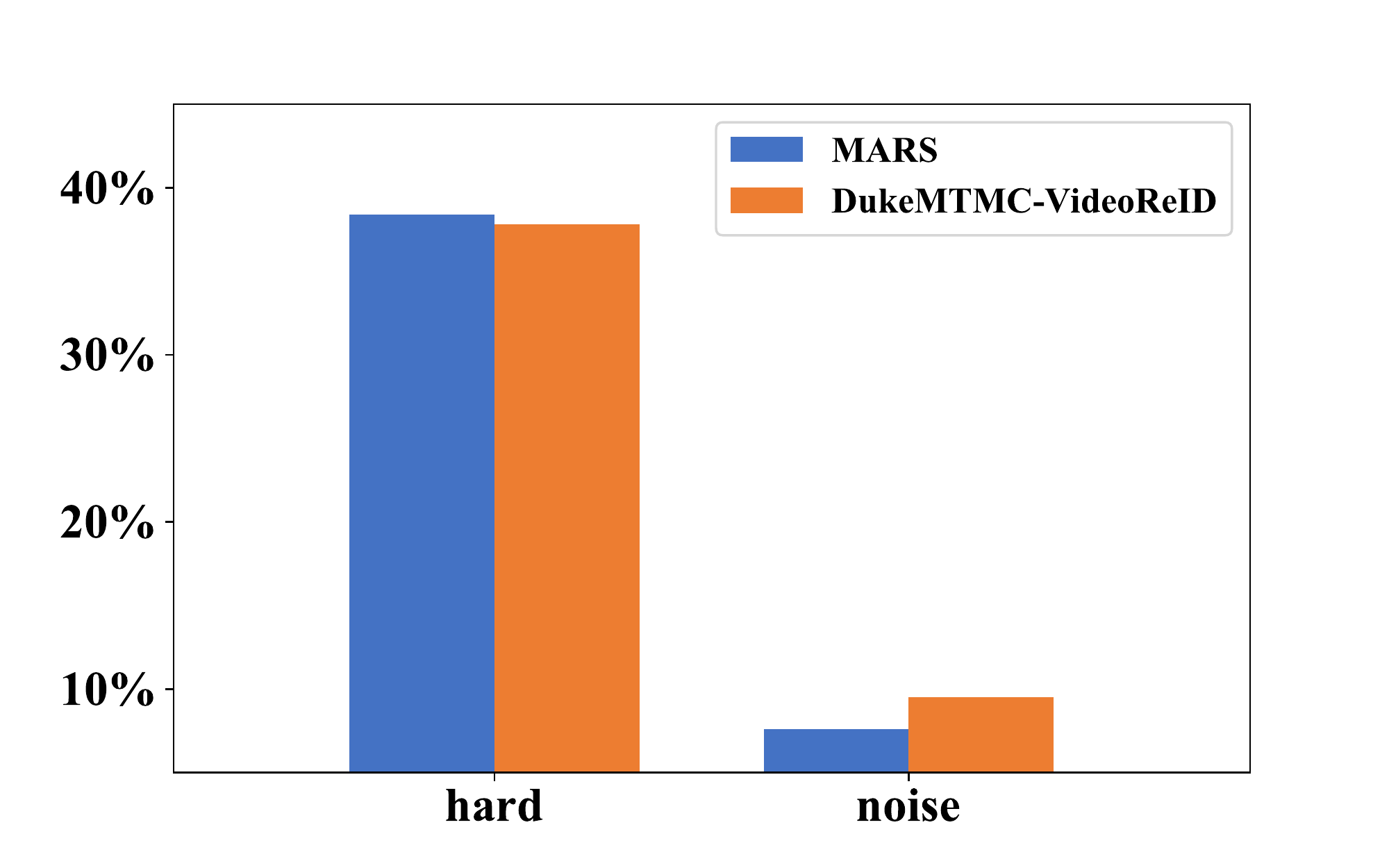}
			}
		\end{tabular}
		\vspace{-4mm}
		\caption{Percentage of hard and noise nodes in the MARS and DukeMTMC-VideoReID datasets.
		}
		\vspace{-5mm}
		\label{fig:ressix}
	\end{figure}
	
	\noindent\textbf{Robustness.}
	Figure \ref{fig:resfour} illustrates the performance change over clustering iterations in the DukeMTMC-VideoReID dataset. Throughout the iteration of our method, the Rank-1 accuracy gradually rises from 44.0\% to 82.8\%; while the mAP accuracy gradually rises from 39.8\% to 76.0\%. After the $14^{th}$ iteration, the model performance stops rising and begins to drop. Compared with the DBC and BUC methods, we observe that our method performs fall behind each other before the $6^{th}$ iteration but diverged afterward, with our outperforming other methods by a relatively large margin. Since our method mines for less varying but more reliable tracklet mergers at the initial stage, this helps our method to merge tracklets across cameras more accurately in the remaining phases.
	Note that, from the $10^{th}$ iteration to the $17^{th}$ iteration, our method always gives the best performance, \textit{i.e.} with a mAP accuracy higher than 67.4\% and a Rank-1 accuracy higher than 75.6\%, which demonstrates the robustness of our proposed method.

	\vspace{-4mm}
	\section{Conclusion}
	We present a Noise and Hard frame Aware Clustering (NHAC) method to handling unsupervised video-based person re-ID tasks. It optimizes both clustering and training steps for the existence of noise and hard frames in tracklets. The graph trimming module suppresses the tracklet feature bias caused by noise nodes. The node re-sampling module enhances the model's learning of rich information from hard nodes. 
	
	Discussion: Figure~\ref{fig:ressix} shows that MARS has a higher percentage of hard nodes and fewer noise nodes compared to DukeMTMC-VideoReID. Since the tracklets in MARS have a higher quality, the improvement of our method on MARS is not so significant as that on DukeMTMC-VideoReID. 
	\vspace{-2mm}
	\section{Acknowledgement}
	This work was supported by the Natural Science Foundation of China (U1803262, 61602349, 61440016, 61801335).
	\vspace{-2mm}
	


\begin{thebibliography}{10}

\bibitem{huang2018video}
Wenjun Huang, Chao Liang, Yi~Yu, Zheng Wang, Weijian Ruan, and Ruimin Hu,
\newblock ``Video-based person re-identification via self paced weighting,''
\newblock in {\em AAAI}, 2018.

\bibitem{wang2020re}
Zheng Wang, Xin Yuan, Toshihiko Yamasaki, Yutian Lin, Xin Xu, and Wenjun Zeng,
\newblock ``Retrieval, verification, and open-set: A new re-identification
  metric,''
\newblock {\em arXiv preprint arXiv:2011.11506}, 2020.

\bibitem{xu2021rethinking}
Xin Xu, Lei Liu, Xiaolong Zhang, Weili Guan, and Ruimin Hu,
\newblock ``Rethinking data collection for person re-identification: active
  redundancy reduction,''
\newblock {\em Pattern Recognition}, vol. 113, pp. 107827, 2021.

\bibitem{wu2018exploit}
Yu~Wu, Yutian Lin, Xuanyi Dong, Yan Yan, Wanli Ouyang, and Yi~Yang,
\newblock ``Exploit the unknown gradually: One-shot video-based person
  re-identification by stepwise learning,''
\newblock in {\em CVPR}, 2018.

\bibitem{ye2017dynamic}
Mang Ye, Andy~J Ma, Liang Zheng, Jiawei Li, and Pong~C Yuen,
\newblock ``Dynamic label graph matching for unsupervised video
  re-identification,''
\newblock in {\em ICCV}, 2017.

\bibitem{liu2017stepwise}
Zimo Liu, Dong Wang, and Huchuan Lu,
\newblock ``Stepwise metric promotion for unsupervised video person
  re-identification,''
\newblock in {\em ICCV}, 2017.

\bibitem{li2020unsupervised}
Minxian {Li}, Xiatian {Zhu}, and Shaogang {Gong},
\newblock ``Unsupervised tracklet person re-identification,''
\newblock {\em TPAMI}, 2020.

\bibitem{chen2018deep}
Yanbei {Chen}, Xiatian {Zhu}, and Shaogang {Gong},
\newblock ``Deep association learning for unsupervised video person
  re-identification,''
\newblock in {\em BMVC}, 2018.

\bibitem{lin2019bottom}
Yutian Lin, Xuanyi Dong, Liang Zheng, Yan Yan, and Yi~Yang,
\newblock ``A bottom-up clustering approach to unsupervised person
  re-identification,''
\newblock in {\em AAAI}, 2019.

\bibitem{ding2019dispersion}
Guodong Ding, Salman~H Khan, Zhenmin Tang, J~Zhang, and F~Porikli,
\newblock ``Dispersion based clustering for unsupervised person
  re-identification,''
\newblock in {\em BMVC}, 2019.

\bibitem{wu2020tracklet}
Guile Wu, Xiatian Zhu, and Shaogang Gong,
\newblock ``Tracklet self-supervised learning for unsupervised person
  re-identification,''
\newblock in {\em AAAI}, 2020.

\bibitem{ye2018robust}
Mang Ye, Xiangyuan Lan, and Pong~C Yuen,
\newblock ``Robust anchor embedding for unsupervised video person
  re-identification in the wild,''
\newblock in {\em ECCV}, 2018.

\bibitem{xiao2017joint}
Tong Xiao, Shuang Li, Bochao Wang, Liang Lin, and Xiaogang Wang,
\newblock ``Joint detection and identification feature learning for person
  search,''
\newblock in {\em CVPR}, 2017.

\bibitem{zheng2016mars}
Liang Zheng, Zhi Bie, Yifan Sun, Jingdong Wang, Chi Su, Shengjin Wang, and
  Qi~Tian,
\newblock ``Mars: A video benchmark for large-scale person re-identification,''
\newblock in {\em ECCV}, 2016.

\end{thebibliography}
\end{document}